%% file: main.tex
\documentclass[sigconf]{acmart}

\usepackage{graphicx} 
\usepackage{subfigure} 
\usepackage{float} 
\usepackage{multirow} 
\usepackage{multicol}
\usepackage{balance} 
\usepackage{amsthm} 

\usepackage{amssymb} 
\usepackage{mathrsfs} 
\usepackage{booktabs}
\usepackage{enumitem}

\copyrightyear{2021} 
\acmYear{2021} 
\setcopyright{acmcopyright}
\acmConference[MM '21]{Proceedings of the 29th ACM International Conference on Multimedia}{October 20--24, 2021}{Virtual Event, China} 
\acmBooktitle{Proceedings of the 29th ACM International Conference on Multimedia (MM '21), October 20--24, 2021, Virtual Event, China} 
\acmPrice{15.00}
\acmDOI{10.1145/3474085.3475583} 
\acmISBN{978-1-4503-8651-7/21/10}

\begin{document}
\fancyhead{}

\title{HetEmotionNet: Two-Stream Heterogeneous Graph Recurrent Neural Network for Multi-modal Emotion Recognition}

\author{Ziyu Jia}
\email{ziyujia@bjtu.edu.cn}
\affiliation{%
  \institution{School of Computer and Information Technology, Beijing Jiaotong University \& Beijing Key Laboratory of Traffic Data Analysis and Mining}
  \city{Beijing}
  \country{China}
}

\author{Youfang Lin}
\email{yflin@bjtu.edu.cn}
\affiliation{%
  \institution{School of Computer and Information Technology, Beijing Jiaotong University \& Beijing Key Laboratory of Traffic Data Analysis and Mining \& CAAC Key Laboratory of Intelligent Passenger Service of Civil Aviation}
  \city{Beijing}
  \country{China}
}

\author{Jing Wang}
\email{wj@bjtu.edu.cn}
\authornote{corresponding author}
\affiliation{%
  \institution{School of Computer and Information Technology, Beijing Jiaotong University \& Beijing Key Laboratory of Traffic Data Analysis and Mining \& CAAC Key Laboratory of Intelligent Passenger Service of Civil Aviation}
  \city{Beijing}
  \country{China}
}

\author{Zhiyang Feng}
\email{zhiyangfeng@bjtu.edu.cn}
\affiliation{%
  \institution{School of Computer and Information Technology, Beijing Jiaotong University}
  \city{Beijing}
  \country{China}
}

\author{Xiangheng Xie}
\email{xiangheng@bjtu.edu.cn}
\affiliation{%
  \institution{School of Computer and Information Technology, Beijing Jiaotong University}
  \city{Beijing}
  \country{China}
}

\author{Caijie Chen}
\email{caijiechen@bjtu.edu.cn}
\affiliation{%
  \institution{School of Computer and Information Technology, Beijing Jiaotong University}
  \city{Beijing}
  \country{China}
}

\begin{abstract}
The research on human emotion under multimedia stimulation based on physiological signals is an emerging field, and important progress has been achieved for emotion recognition based on multi-modal signals. However, it is challenging to make full use of the complementarity among spatial-spectral-temporal domain features for emotion recognition, as well as model the heterogeneity and correlation among multi-modal signals. In this paper, we propose a novel two-stream heterogeneous graph recurrent neural network, named HetEmotionNet, fusing multi-modal physiological signals for emotion recognition. Specifically, HetEmotionNet consists of the spatial-temporal stream and the spatial-spectral stream, which can fuse spatial-spectral-temporal domain features in a unified framework. Each stream is composed of the graph transformer network for modeling the heterogeneity, the graph convolutional network for modeling the correlation, and the gated recurrent unit for capturing the temporal domain or spectral domain dependency. Extensive experiments on two real-world datasets demonstrate that our proposed model achieves better performance than state-of-the-art baselines.
\end{abstract}



\maketitle

\section{INTRODUCTION}

Emotion is a mental and physiological state which results from many senses and thoughts \cite{james2013emotion}. In recent years, emotion plays an increasingly important role in multiple areas, such as disease detection, human-computer interaction (HCI), and virtual reality. To collect data for emotion recognition, researchers often utilize multiple multimedia materials to stimulate participants and induce emotion. For the reason that videos, audio, and so on have the advantage of easy collection, there is a mass of studies using these data. Though they achieve certain performance, these data cannot guarantee to reflect a real emotional state since human can disguise their facial expression, sound, and so on. Physiological signals cannot be disguised, so they can reflect human emotion objectively. Therefore, physiological signals are more suitable for emotion recognition.

\begin{figure}[htb]
\setlength{\abovecaptionskip}{0cm}
\setlength{\belowcaptionskip}{-0.5cm}
\centering
\includegraphics[width=1\linewidth]{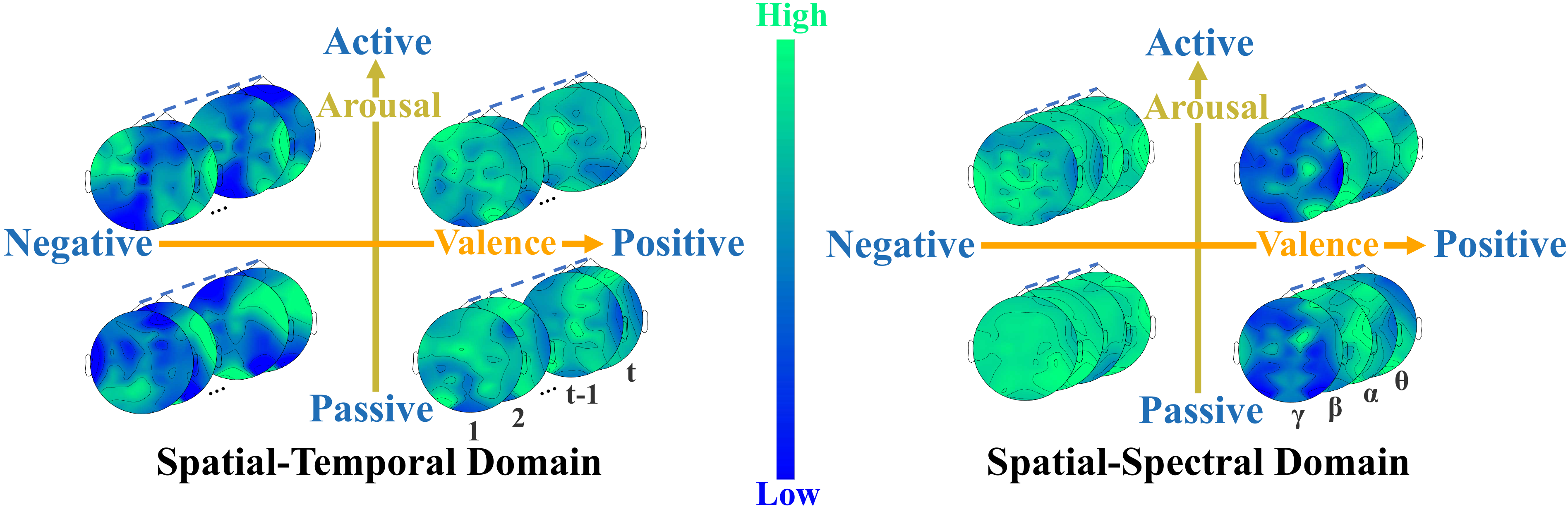}
\caption{The complementarity among spatial-spectral-temporal domain features. The brain topographic maps indicate that there are differences of activation degree between temporal domain information and spectral domain information on the spatial domain for different emotional states which are categorized by valence and arousal. These differences in spatial-spectral-temporal domain information are complementary for emotion recognition.}
\label{intro_fig1}
\end{figure} 

\begin{figure}[htb]
\setlength{\abovecaptionskip}{0cm}
\setlength{\belowcaptionskip}{-0.5cm}
\centering
\includegraphics[width=0.8\linewidth]{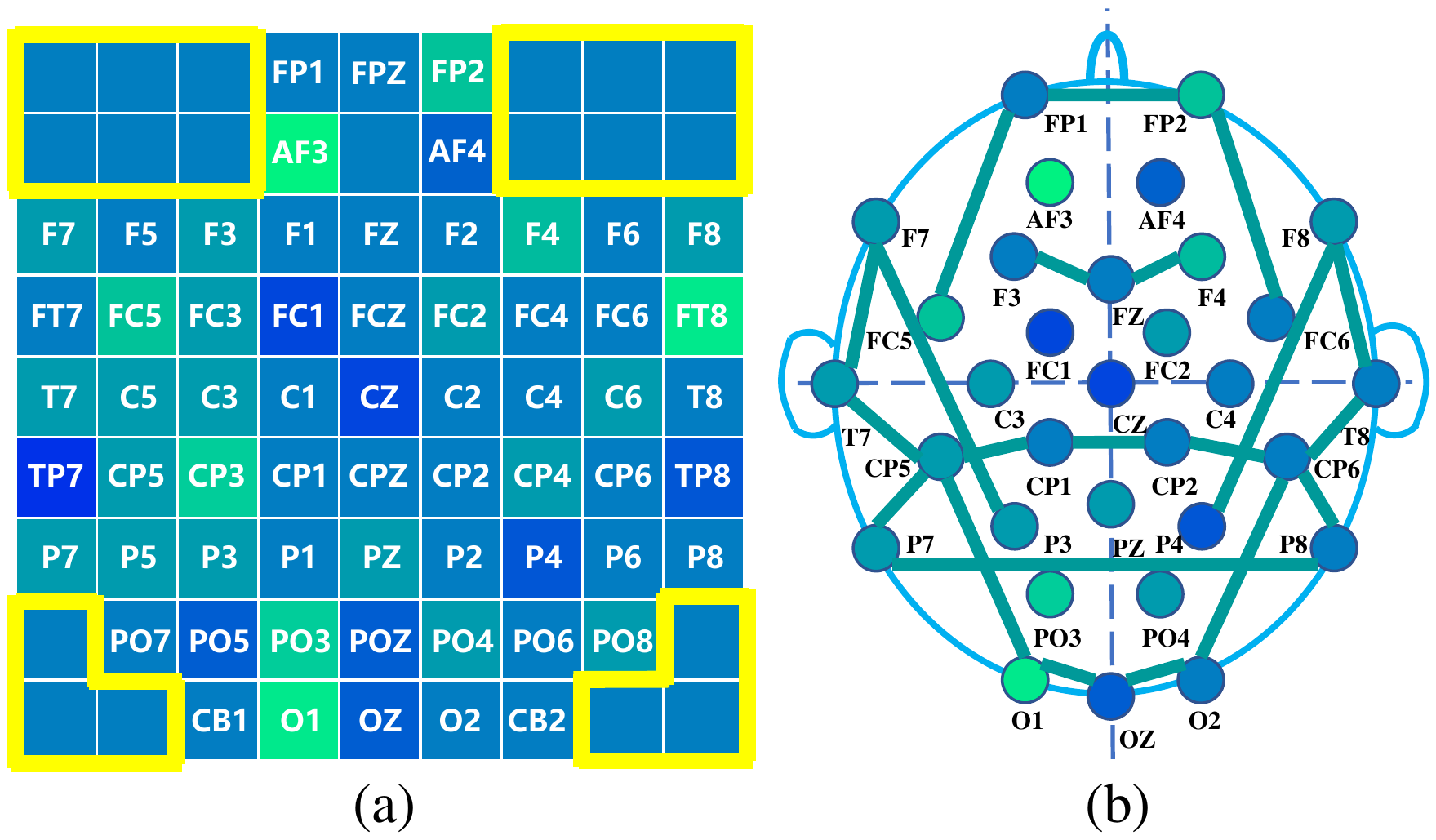}
\caption{Two different EEG representations. (a) The 2D map of EEG signals. The EEG signals are constructed into image-like data and CNN is applied to extract features from these data. However, those marginal areas circled by the yellow border are often filled with zero, which may introduce noise and limit the performance of CNN. Because the voltage of these areas not covered by biosensors is usually not zero, artificial zero-filling may introduce wrong information. (b) The graph of EEG signals. The graph method preserves the functional connectivity among the electrodes to the greatest extent and does not introduce noise.}
\label{intro_fig2}
\end{figure}
  
To achieve better performance for emotion recognition, some methods are proposed which use physiological signals and achieve state-of-the-art performance \cite{ma2019emotion,tao2020eeg,jia2020sst}. However, several challenges still exist for emotion recognition using physiological signals.

\textbf{C1: How to utilize the complementarity among spatial-spectral-temporal domain information efficiently.} Temporal domain information and spectral domain information in the spatial domain of physiological signals usually have different activation degree. For example, Figure \ref{intro_fig1} indicates the differences between the temporal domain and the spectral domain in the spatial domain of electroencephalography (EEG) signals for different emotional states. To be specific, in the spatial-temporal domain, the activation degree of the temporal domain information directly reflects brain activity. The high degree of activation is usually related to positive emotion and the low degree of activation is often related to negative emotion. In the spatial-spectral domain, the activation degree of the $\gamma$ band is often high in negative emotions and low in positive emotions \cite{martini2012dynamics}. Considering these phenomena, most studies design various deep learning methods to extract spatial-temporal domain information or spatial-spectral domain information for emotion recognition. These methods can be mainly divided into two categories. One concentrates on the spatial-temporal domain \cite{liao2020multimodal,salama2018eeg,wen2017novel,dar2020cnn}, and the other one concentrates on the spatial-spectral domain \cite{zhao2019multimodal,yang2018recurrence,yang2018continuous,zheng2018emotionmeter}. Specifically, the methods focusing on spatial-temporal domain information often only take the raw signals or other statistical features as the input of the network. 
The methods focusing on the spatial-spectral domain often capture the correlation of the spectral domain features such as power spectral density (PSD) and differential entropy (DE) in the physiological signals. 
Though these two categories of methods achieve relatively satisfactory performance, they ignore complementarity among spatial-spectral-temporal domain information. SST-EmotionNet is proposed to extract the spatial-spectral-temporal domain information in the image-like maps \cite{jia2020sst}. However, areas of brain that are not covered by biosensors also produce bioelectrical signals. The voltage of these signals is usually not 0, so artificial zero-filling may introduce noise, as shown in Figure \ref{intro_fig2}(a). Besides, in Figure \ref{intro_fig2}(b), the graph-based method can reflect the topological relationship of the brain \cite{wu2020comprehensive, jia2020graphsleepnet}. These shortcomings limit the performance of classification to a certain extent. Therefore, how to effectively extract spatial-spectral-temporal domain information is challenging work. 

\begin{figure}[htb]
\setlength{\abovecaptionskip}{0cm}
\setlength{\belowcaptionskip}{-0.5cm}
  \centering
  \includegraphics[width=1\linewidth]{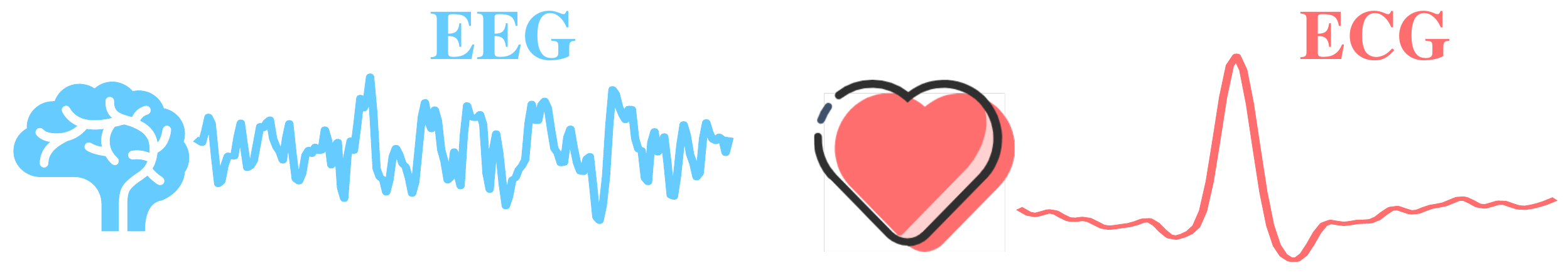}
  \caption{The heterogeneity of the multi-modal physiological signals. For instance, there are differences in the waveform and amplitude between ECG and EEG signals.}
  \label{intro_fig3}
  \end{figure}

\textbf{C2: How to model heterogeneity and correlation among different modalities simultaneously.} The heterogeneity and the correlation exist in multi-modal physiological signals. The heterogeneity among modalities is reflected in the differences among the attributes of various signals, which are collected from different organs \cite{jiang2020snapshot}. For instance, electrocardiography (ECG) and EEG signals have a great difference in waveform and amplitude, as shown in Figure \ref{intro_fig3}. The correlation includes intra-modality correlation and cross-modality correlation. The intra-modality correlation is the relationship among channels in the same modality, such as the functional connectivity as shown in Figure \ref{intro_fig2}(b). The cross-modality correlation is the relationship among channels in different modalities. For example, when participants are in fear, a greater heart rate acceleration is reflected in ECG signals, accompanied by an increase in GSR signals, and high activation degree of EEG signals in the right frontal lobe \cite{lichtenstein2008comparing, kreibig2010autonomic}. However, most current methods only model the heterogeneity or correlation among different modalities separately. As for the heterogeneity, existing works use different feature extractors, such as CNN, to extract features of different modalities and capture the heterogeneity in multi-modal data \cite{mittal2020m3er, liu2019multimodal, ma2019emotion, lin2017deep}. However, these methods ignore the correlation among different modalities because the signals of different modalities are extracted separately. As for the correlation, existing works usually combine the signals of different modalities into a new data representation and feed it to a deep neural network to capture the correlation \cite{pinto2019biosignal, liu2016multimodal}, such as group sparse canonical correlation analysis (GSCCA) \cite{zhang2020fusing}. However, these methods ignore the heterogeneity among different modalities because the features of all modalities are extracted by the same feature extractor. Therefore, how to model the heterogeneity and correlation among multi-modal signals simultaneously becomes a challenge.
 
To address these challenges, we propose a two-stream heterogeneous graph recurrent neural network named HetEmotionNet, which is composed of the spatial-temporal stream and spatial-spectral stream and takes heterogeneous graph sequences as input. A stream consists of a graph transformer network (GTN), a graph convolutional network (GCN), and a gated recurrent unit (GRU). The physiological signals are constructed into heterogeneous graph sequences as the input of our model. Several heterogeneous graphs are stacked to form a heterogeneous graph sequence. A heterogeneous graph consists of nodes and edges. A node represents a channel of signals whose type depends on the modality of the channel. An edge represents the connection between two nodes.
To sum up, the main contributions of our work are summarized as follows:

$\bullet$ We propose a novel graph-based two-stream structure composed of the spatial-temporal stream and the spatial-spectral stream which can simultaneously fuse spatial-spectral-temporal domain features of physiological signals in a unified deep neural network framework.

$\bullet$ Each stream consists of a GTN for modeling the heterogeneity, a GCN for modeling the correlation, and a GRU for capturing the temporal or spectral dependency.

$\bullet$ Extensive experiments are conducted on two benchmark datasets to evaluate the performance of the proposed model. The results indicate the proposed model outperforms all the state-of-the-art models.

\section{RELATED WORK}

\begin{figure*}[htb]
\setlength{\abovecaptionskip}{-0.2cm}
\setlength{\belowcaptionskip}{-0.5cm}
\centering
\includegraphics[width=1\linewidth]{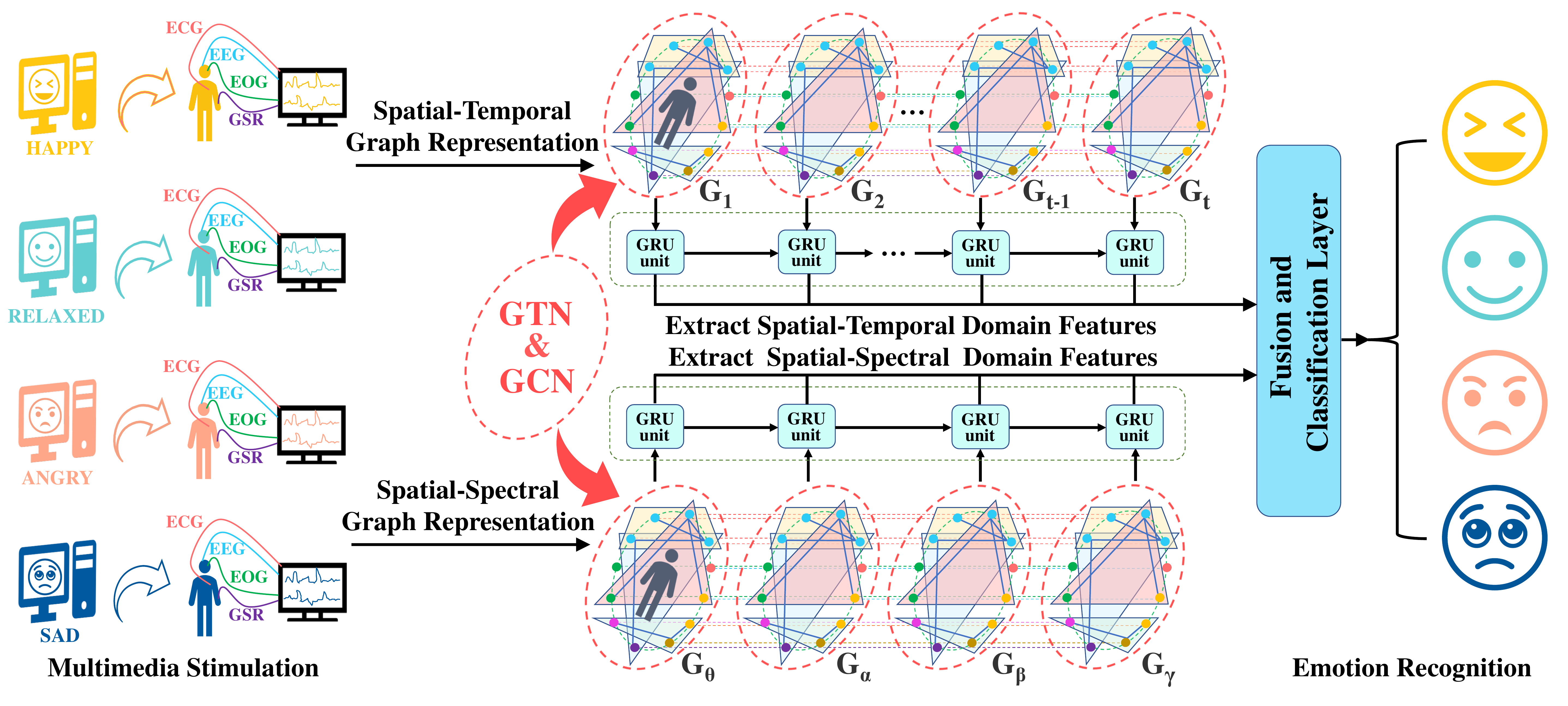}
\caption{The whole schematic process for multi-modal emotion recognition. We obtain the multi-modal signals from the subjects and construct the graph sequence representation for the HetEmotionNet. HetEmotionNet consists of spatial-temporal stream and spatial-spectral stream. The two streams have the same structure. Each stream is composed of graph transformer network (GTN), graph convolutional network (GCN), and gated recurrent unit (GRU).}
\label{graph_sequence_generation}
\end{figure*}

In recent years, time series analysis has attracted the attention of many researchers \cite{jia2020refined,jia2019detecting}. As a typical time series, physiological signals are used in many fields, such as sleep staging \cite{jia2020graphsleepnet,cai2020brainsleepnet,jia2021salientsleepnet,jia2020sleepprintnet,jia2020sleep}, motor imagery \cite{li2020learning,ziyu2020motor,10.1007/978-3-030-67664-3_44}, seizure detection \cite{liu2020representation,liu2021new}, and emotion recognition \cite{jia2020sst,zheng2018emotionmeter,schirmer2017emotion}, etc. The physiological signals have been applied in emotion recognition widely because they can accurately reflect real emotions. Musha et al. use EEG signals to recognize emotions for the first time \cite{musha1997feature}. In the early stage of research, researchers adopted traditional machine learning models like SVM to extract features of EEG signals to recognize emotions \cite{lin2005speech}. However, those methods are often limited by prior knowledge. Deep learning methods \cite{zheng2018emotionmeter,schirmer2017emotion,lu2015combining,zheng2017identifying,song2018eeg,tao2020eeg} are proposed to make up for the shortcomings of traditional machine learning methods. They have shown outstanding results in natural language processing \cite{bahdanau2014neural} and computer vision \cite{simonyan2014very} as well as in affective computing. 

\textbf{Multi-domain Features Extraction.} In building deep learning models for emotion recognition, researchers often extract spatial-temporal domain features or spatial-spectral domain features to train models. For instance, for \textbf{spatial-temporal domain features} extraction, Zhang et al. propose a spatial-temporal recurrent neural network combining the spatial RNN layer and the temporal RNN layer to capture the spatial and temporal features, respectively \cite{zhang2018spatial}. Yang et al. propose a parallel convolutional recurrent neural network, which utilizes CNN to extract spatial features from data frames and RNN to extract temporal features from EEG signals \cite{yang2018emotion}. For \textbf{spatial-spectral domain features} extraction, Yang et al. propose a 3D EEG representation, which combines features from different frequency bands while retaining spatial information among channels \cite{yang2018continuous}. Zheng et al. utilize Short-Time Fourier Transform with a non-overlapped Hanning window to extract features from multi-channel EEG signals and calculate DE features for each frequency band to predict the positive and negative emotional states \cite{zheng2014eeg}. Further, Jia et al. present SST-EmotionNet, extracting spatial features, spectral features, and temporal features simultaneously, and achieving higher performance for emotion recognition. However, image-like maps introduce noise into the network \cite{jia2020sst}.

\textbf{Multi-modal Physiological Signals.} Most methods model either heterogeneity \cite{mittal2020m3er, liu2019multimodal} or correlation \cite{pinto2019biosignal} to fuse multi-modal signals. For instance, for the \textbf{heterogeneity of multi-modal data}, Lin et al. extract the features of EEG signals and Peripheral Physiological Signals (PPS) separately, and then concatenate these features to model the heterogeneity of multi-modal signals for emotion recognition \cite{lin2017deep}. Ma et al. design a multi-modal residual LSTM network (MMResLSTM) for emotion recognition. The signals of different modalities are fed into different LSTM branches to extract multi-modal features \cite{ma2019emotion}. For the \textbf{correlation of multi-modal data}, Zhang et al. adopt GSCCA to learn the correlation by extracting the group structure information between EEG signals and eye movement features \cite{zhang2020fusing}. Liu et al. fuse features of EEG signals and features of other modalities with bimodal deep autoencoders and then obtain the share representation \cite{liu2016multimodal}.
  
Overall, existing emotion recognition methods have achieved high accuracy. However, they still ignore the fusion of multi-domain features, multi-modal heterogeneity, and correlation. To solve the limitations, we propose HetEmotionNet, which models the complementarity among spatial-spectral-temporal domain features, the heterogeneity, and the correlation of the signals simultaneously.

\section{PRELIMINARIES}

\quad\textbf{Definition 1. Heterogeneous Graph.} We define $G=(V,E)$ as a graph, where $V$ denotes the node set and $E$ denotes the edge set. The network schema is defined as $ \mathcal{T}_{G} = (\mathcal{A},\mathcal{R}) $, where $\mathcal{A}$ is the node type set and $\mathcal{R}$ is the edge type set. We define $\phi: V \rightarrow \mathcal{A}$ as the node type mapping and $\psi: E \rightarrow \mathcal{R}$ as the edge type mapping. Therefore, a network can be defined as $ G = (V,E,\phi,\psi) $. For a heterogeneous network $ G = (V,E,\phi,\psi) $, |$\mathcal{A}$| + |$\mathcal{R}$| > 2. 

\textbf{Definition 2. Heterogeneous Emotional Network.}
The heterogeneous emotional network $ G^{ER} = (V^{'}, E^{'}, \phi^{'}, \psi^{'}) $ is constructed based on multi-modal signals. We define $V^{'}=(v_1, v_2, \dots, v_N)$ as the node set, where the node $v_i$ denotes a channel of physiological signals. The edge represents a connection relationship between two different channels. $ E^{'} = (e_1, e_2, \dots, e_{n}) $ is defined as the edge set, where $n$ is the number of edges. The number of node types is equal to the number of modalities. When the number of modalities is $A^{'}$, the number of edge types is $ R^{'} = A^{'}(A^{'}+1)/2 $. After the node type mapping $\phi^{'}: V^{'} \rightarrow A^{'}$ and the edge type mapping $\psi^{'}:E^{'} \rightarrow R^{'}$ are defined, we construct the heterogeneous emotional network $G^{ER} = (V^{'}, E^{'}, \phi^{'}, \psi^{'})$. In addition, the heterogeneous emotional network can be defined as $G^{ER} = (X^F, A)$, where $X^F$ denotes the features of each node and $A$ denotes the adjacency matrix of $G^{ER}$.

\textbf{Definition 3. Heterogeneous Graph Sequence.} We define $G_s = (G_1, G_2, \dots,G_{n})$ as a heterogeneous graph sequence, where $n$ is the number of heterogeneous graphs. For instance, we construct the spatial-temporal graph sequence $G^{T} = (G_1, G_2, \dots, G_T) \in\mathbb{R}^{N\times T}$, where $T$ denotes the number of the timesteps in the sample. Analogously, we construct the spatial-spectral graph sequence $G^{B} = (G_1, G_2, \dots, G_B)  \in\mathbb{R}^{N\times B}$, where $B$ is the number of frequency bands.

\noindent \textbf{Problem Statement.} Our research goal is to learn a mapping function between graph sequences and emotional states. Given spatial-temporal graph sequence $G^{T}$ and spatial-spectral graph sequence $G^{B}$, the emotion recognition problem can be defined as $ Y = F(\mathbf{G}^{T}, \mathbf{G}^{B}) $, where $Y$ denotes the emotion classification and $F$ denotes the mapping function.

\section{METHODOLOGY}
Figure \ref{graph_sequence_generation} illustrates the overall structure of our model. We propose a HetEmotionNet based on the constructed graph sequence from the multi-modal signals. HetEmotionNet consists of a spatial-temporal stream and a spatial-spectral stream, and these two streams have the similar structure. Each stream of HetEmotionNet is composed of graph transformer network, graph convolutional network, and gated recurrent unit. We summarize five core ideas of HetEmotionNet: 1$)$ Design a heterogeneous graph-based spatial-spectral-temporal representation for multi-modal emotion recognition. 2$)$ Integrate the graph-based spatial-temporal stream and spatial-spectral stream in a unified network structure to extract spatial-spectral-temporal domain features simultaneously. 3$)$ Adopt graph transformer network to model the heterogeneity of multi-modal signals. 4$)$ Apply graph convolutional network to capture the correlation among different channels. 5$)$ Employ the gated recurrent unit to capture temporal domain or spectral domain dependency, respectively.
\begin{figure}[htb]
\setlength{\abovecaptionskip}{0cm}
\setlength{\belowcaptionskip}{-0.5cm}
  \centering
  \includegraphics[width=1\linewidth]{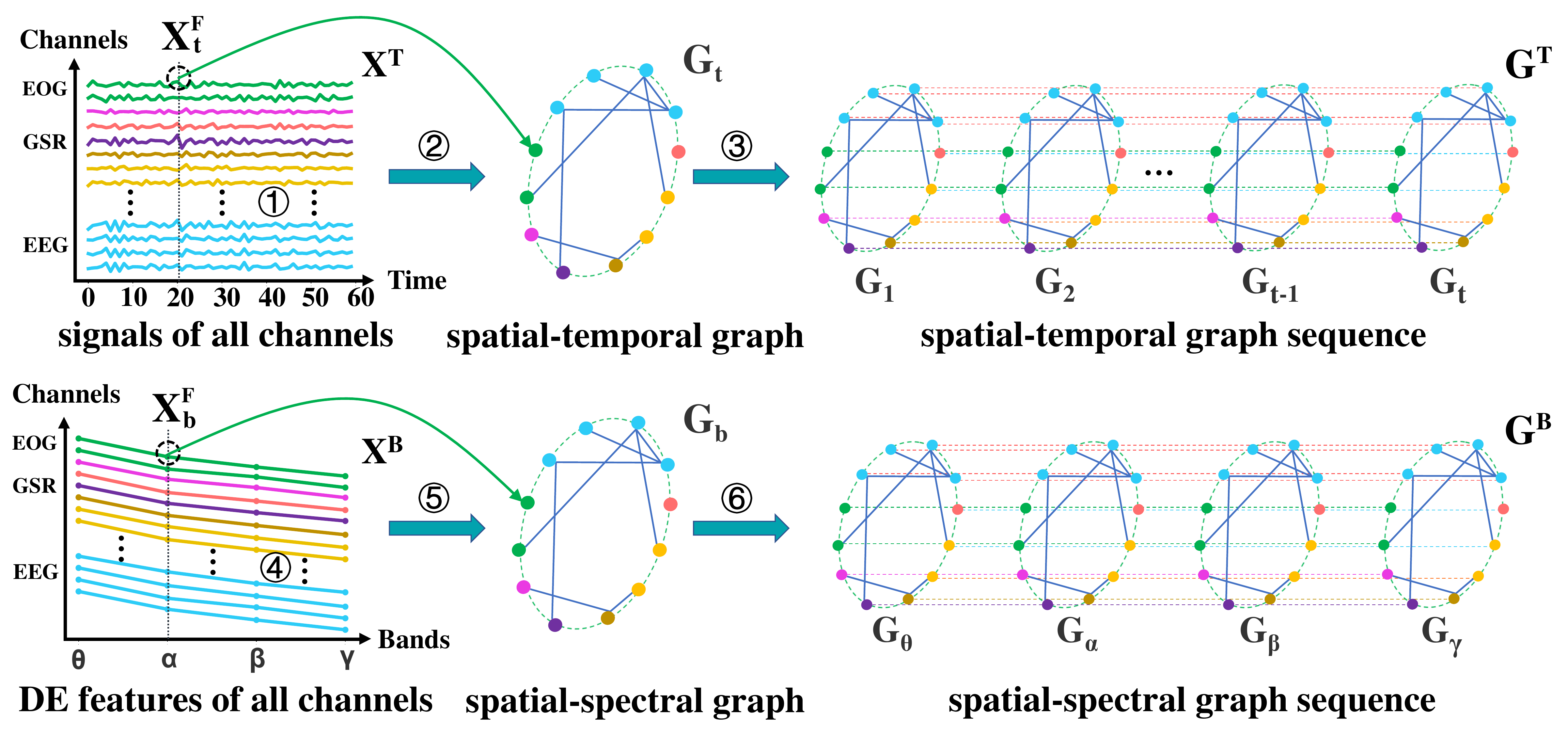}
  \caption{The schematic process of heterogeneous graph sequence construction. The heterogeneous graph sequence is stacked by several heterogeneous graphs. A heterogeneous graph is composed of the node features and graph structure. Each node corresponds to a channel, whose node type depends on the modality of channel. Graph structure represents the relationship among channels.}
  \label{graph_sequence_construction}
  \end{figure}

\subsection{Heterogeneous Graph Sequence Construction}\label{Graph Sequence Construction}
For each sample, we construct a heterogeneous spatial-temporal graph sequence and a heterogeneous spatial-spectral graph sequence separately as shown in Figure \ref{graph_sequence_construction}. These heterogeneous graph sequences are used to describe the spatial distribution of the temporal domain and spectral domain information of multi-modal signals. The heterogeneous graph sequence is made up of several heterogeneous graphs. The general heterogeneous graph is usually defined as $G=(X^F,A)$, where $X^F$ denotes the graph node features, $A$ denotes the graph adjacency matrix which describes the relationship of channels. Besides, the node type depends on the modality of the channel.

Figure \ref{graph_sequence_construction} presents the process of constructing the heterogeneous graph sequence. Firstly, we build the relationship of the channels as \textcircled{1} in Figure \ref{graph_sequence_construction}. Specifically, we calculate the correlation degree among different channels in the sample to determine the relationship of the channels using mutual information \cite{kraskov2004estimating}. Formally, given a channel pair $(u,v)$, their correlation $a_{u,v}$ can be formulated as follows:
\begin{equation}
a_{u,v}=I(X^T_u;X^T_v)=\sum_{m\in X^T_u}\sum_{n\in X^T_v}p(m,n)\log{\frac{p(m,n)}{p(m)p(n)}},
\end{equation}
where $1 \le u \le N$, $1 \le v \le N$, $X_u^T$ represents the signals of channel $u$, $X_v^T$ represents the signals of channel $v$, $T$ is the number of timesteps, and $N$ is the number of channels. After calculating the correlation degree for all the channel pairs, we obtain the adjacency matrix $A=(a_{1,1}, \dots, a_{u,v}, \dots, a_{N,N})\in \mathbb{R}^{N\times N}$ of the spatial-temporal graphs and the spatial-spectral graphs.

To construct the heterogeneous spatial-temporal graph sequence, we compute the temporal node features $X^F_t=(X^{F1}_t, X^{F2}_t, \dots, X^{FN}_t)\in \mathbb{R}^N$, where $X^{Fi}_t$ denotes the amplitude ($\mu V$) on the $i^{th}$ channel in the $t^{th}$ timestep from the multi-modal signals, $1\le t\le T$. Then, we combine the temporal domain feature vector $X^F_t$ and the adjacency matrix $A$ to form a heterogeneous spatial-temporal graph $G_t=(X^F_t,A)$ as \textcircled{2}. Therefore, we obtain $T$ heterogeneous spatial-temporal graphs from the sample. These heterogeneous spatial-temporal graphs are stacked into a heterogeneous spatial-temporal graph sequence $G^T=(G_1,G_2,\dots,G_{T-1},G_T )$ as \textcircled{3}.

To construct the heterogeneous spatial-spectral graph sequence, similar to the process of generating heterogeneous spatial-temporal graph sequence, we extract the DE features from the $B\in Z^+$ frequency bands in the sample as \textcircled{4}. These features extracted from different channels and the correlation among channels are converted to heterogeneous spatial-spectral graphs. Specifically, these feature vectors extracted from $N$ channels in the $b^{th}$ frequency band $X^F_b=(X^{F1}_b, X^{F2}_b, \dots, X^{FN}_b)\in \mathbb{R}^N$ and the adjacency matrix $A$ of the sample are composed of a heterogeneous spatial-spectral graph $G_b=(X^F_b,A)$ as \textcircled{5}. Therefore, we obtain $B$ heterogeneous spatial-spectral graphs in $B$ frequency bands for each sample. Then, these heterogeneous spatial-spectral graphs are stacked into a heterogeneous spatial-spectral sequence $G^B=(G_\theta,G_\alpha,G_\beta,G_\gamma)$ as \textcircled{6}.

\subsection{Heterogeneous Graph Recurrent Neural Network}
HetEmotionNet consists of graph transformer network and graph recurrent neural network. Graph recurrent neural network is composed of graph convolutional network and gated recurrent unit.

\textbf{Graph Transformer Network.}
The Graph Transformer Network (GTN) \cite{yun2019graph} is used to model the heterogeneity of multi-modal signals by automatically extracting the meta-paths from the adjacency matrix set $\mathcal{A}$. Specifically, $\mathcal{A}$ is generated from the heterogeneous adjacency matrix obtained in Section \ref{Graph Sequence Construction}. The heterogeneous adjacency matrix is divided into homogeneous adjacency matrix set $\mathcal{A} = \{\mathcal{A}_1, \mathcal{A}_2, \dots, \mathcal{A}_{N_a}\}$ according to edge types, where $\mathcal{A}_i$ is homogeneous adjacency matrix, $1 \le i \le N_a$, and $N_a$ is the number of adjacency matrices which is decided by the number of edge types.

GTN consists of several GT-layers. Each GT-layer aims to learn a two-level meta-path, which represents the second-order neighbor relationship in the graph by learning two graph structures from $\mathcal{A}$. GT-layer has two steps: the first step is designed to softly construct two graph structures, which are defined as:
\begin{equation}
Q_1 = \phi(\mathcal{A},softmax(W_{\phi1})),
\end{equation}
\begin{equation}
Q_2 = \phi(\mathcal{A},softmax(W_{\phi2})),
\end{equation}where $\phi$ denotes the $1 \times 1$ convolution, $W_{\phi1} \in \mathbb{R}^{1 \times 1 \times K}$ and $W_{\phi2}  \in \mathbb{R}^{1 \times 1 \times K}$ are the parameters of $\phi$. $Q_1$ and $Q_2$ are the graph structures constructed by the $1 \times 1$ convolution.

The second step is designed to generate meta-paths by matrix multiplication of $Q_1$ and $Q_2$. Formally, $H^{(l)} = Q_1Q_2$, where $H^{(l)}$ represents the meta-paths generated by $l^{th}$ GT-layer. For calculation stability, $H^{(l)}$ is updated as $H^{(l)} = D^{-1}Q_1Q_2$, where $D$ denotes the degree matrices of $\mathcal{A}$.

Stacking several GT-layers in GTN aims to learn a high-level meta-path that is a useful relationship of multi-modal signals. The $l^{th}$ GT-layer in GTN updates the graph structure and multiplies the graph structure with the one constructed by $(l - 1)^{th}$ GT-layer. Formally, $H^{(l)} = Q_lH^{(l - 1)}$, where $Q_l$ denotes the graph structure learned by $l^{th}$ GT-layer, $H^{(l)}$ and $H^{(l - 1)}$ represent the meta-paths learned by $(l - 1)^{th}$ and $l^{th}$ GT-layers, respectively. To consider multiple relationship of multi-modal signals, which are decided by the number of meta-paths simultaneously, multi-channel $1 \times 1$ convolution is applied. The graph structure generated by multi-channel $1 \times 1$ convolution is defined as:
\begin{equation}
Q_l^{'} = \Phi(\mathcal{A}, softmax(W_{\Phi})),
\end{equation}where $\Phi$ is the multi-channel $1 \times 1$ convolution, $W_{\Phi} \in \mathbb{R}^{1 \times C^A \times K}$ is learnable parameter, $C^A$ is the number of channels, and $Q_l^{'}$ is the graph structure generated from $l^{th}$ GT-layer by this convolution. To concurrently learn various lengths of meta-paths, the identity matrix is added into the result of each GT-layer. Finally, the graph sequence is updated into $C^A$ sequences according to the meta-paths learned from GTN.

\textbf{Graph Recurrent Neural Network.}
Graph recurrent neural network has two chief components, namely graph convolutional network and gated recurrent unit. These two components are used to extract both spatial domain and temporal/spectral domain features from graph sequences. Specifically, GCN is used to capture the spatial domain features by aggregating information from neighbor nodes. In addition, GRU is applied to extract temporal/spectral domain features from the graph sequence obtained after GCN.

\emph{Graph Convolutional Network.}
GCN captures the correlation among the nodes in the graph. The correlation is valuable to recognize different emotions. For example, there exists much functional connectivity among different brain regions. Capturing the spatial domain relationship among these functional connectivity contributes to recognizing different emotions.

To capture the correlation, the spectral graph theory extends the convolution operation from image-like data to graph structure data \cite{bruna2013spectral}. The graph structure is represented by its Laplacian matrix in the spectral graph analysis. The Laplacian matrix and its eigenvalues reflect the property of the graph. The Laplacian matrix is defined as $L = D - A$, where $L \in \mathbb{R}^{N \times N}$ is the Laplacian matrix, $A \in\mathbb{R} ^{N \times N}$ is the adjacency matrix, $D \in \mathbb{R}^{N \times N}$ is the degree matrix of the graph, and $N$ is the vertex number. The eigenvalues matrix is obtained by decomposing the Laplacian matrix with $L = U \Lambda U^T$, where $U$ is the Fourier basis and $\Lambda \in \mathbb{R}^{N \times N}$ is a diagonal matrix. The node feature is defined as $x_i \in \mathbb{R}^{N}$ for the graph $G_i$ in graph sequence $G_s$. Then, the graph Fourier transform and inverse Fourier transform are defined as $\hat{x_i} = U^T x_i$ and $x_i = U\hat{x_i}$, respectively. Because the graph convolution operation is equal to the product of these signals which have been transformed into the spectral domain \cite{bruna2013spectral}, the graph convolution of signals $x_i$ on graph $G_i$ is defined as:\begin{equation}
    g_\theta \star_{G_i}  x_i  = U g_\theta U^T x_i,
\end{equation}where $g_\theta$ denotes a graph convolution kernel and $\star_{G_i}$ denotes a graph convolution operation on graph $G_i$. In addition, to simplify the calculation of Laplace matrix, 
the $K$ order Chebyshev Polynomials are adopted, which is defined as:\begin{equation}
    g_{\theta}(\Lambda) \approx \sum_{k=0}^{K} \theta_{k} T_{k}(\tilde{\Lambda}),
\end{equation}where $\tilde{\Lambda} = \frac{2}{\lambda_{max}} \Lambda - I_N$, $I_N$ is the identity matrix, and $\lambda_{max}$ is the maximum eigenvalues of $\Lambda$. The $\theta \in R^{K}$ is a vector of polynomial coefficients. The Chebyshev polynomials are defined as $T_k(x) = 2 x T_{k - 1}(x) - T_{k - 2}(x)$ recursively, where $T_0(x) = 1$ and $T_1(x) = x$. With the $K$ order Chebyshev Polynomials, the message from 0 to $K^{th}$ neighbours are aggregated into the center node. For the fast calculation \cite{kipf2016semi}, the first-order polynomials are used and $\lambda_{max} = 2$, $k = 1$ are set. Then, the graph convolution is further simplified as:
\begin{equation}
    g_{\theta} \star_{G_i} x_i \approx \theta\left(I_{N}+D^{-\frac{1}{2}} A D^{-\frac{1}{2}}\right) x_i,
\end{equation}
where $D$ is the degree matrix, $A$ is the adjacency matrix, and $\theta$ is the parameter. $I_{N}+D^{-\frac{1}{2}}AD^{-\frac{1}{2}}$ is changed into $\tilde{D}^{-\frac{1}{2}} \tilde{A} \tilde{D}^{-\frac{1}{2}}$ to prevent numerical instability, gradient explosion and gradient disappearance, where $\tilde{D} = \sum_j \tilde{A}_{i j}$ and $\tilde{A} = A + I_N$. The graph convolution of signals $x_i$ on graph $G_i$ in graph sequence is defined as:
\begin{equation}
    H_i^{GCN}=\sigma \left(\tilde{D}^{-\frac{1}{2}} \tilde{A} \tilde{D}^{-\frac{1}{2}} x_i W\right) + b,
\end{equation}
where $x_i$ and $H_i^{GCN}$ are the node features before and after the graph convolution, respectively. $W$ is the weight vector, $\sigma$ is the Rectified Linear Unit (ReLU) activation function, and $b$ is a bias. 

The graph convolution operation is applied to the graph sequence. Formally, for graph sequence $G_s = \{G_1, G_2, \dots,G_{ns}\} \in \mathbb{R}^{N \times ns}$ and each graph $G_i$ in $G_s$, the graph convolution operation is applied on $G_i$. After the graph convolution operation is performed on all graphs in $G_s$, the results are stacked together to construct a new graph sequence $H_s = \{H_1, H_2, \dots, H_{ns}\} \in \mathbb{R}^{N \times ns}$. To decrease the number of parameters, all graph convolution operations share the same parameters.

Since the meta-paths learned by GTN update the graph sequence obtained in Section 4.2 into $C^A$ graph sequences, the graph convolution operation is performed on each of the graph sequences. Then, the weighted sum operation is adopted to concatenate the graph sequences together. The weighted sum operation on graph sequences $G^{GTN}_s$ is defined as:\begin{equation}
    H_s^{GCN} = \sum_{i}^{C^A} w_i \delta_s(G_{s_i}^{GTN}),
\end{equation}where $\delta_s$ denotes the graph convolution operation on the graph sequence, $w_i$ is the weight for the weighted sum operation, and $G_{s_i}^{GTN}$ is the $i^{th}$ graph sequence in $G_s^{GTN}$.

\emph{Gated Recurrent Unit.}
After the graph convolutional network, GRU is applied to capture the dependency among different timesteps or frequency bands. Gated recurrent unit consists of the reset gate and the update gate \cite{chung2014empirical}, which is defined as:
\begin{equation}
R_{i}=\sigma\left(W_{r} H_{i}^{GCN}+V_{r} H_{i-1}^{GRU}\right)+b_{r},
\end{equation}
\begin{equation}
\tilde{H}_{i}^{GCN}= \tanh \left(W_{h} H_{i}^{GCN}+V_{h}\left(R_{i} \odot H_{i-1}^{GRU}\right)+b_{h}\right),
\end{equation}
\begin{equation}
Z_{i}=\sigma\left(W_{z} H_{i}^{GCN}+V_{z} H_{i-1}^{GRU}\right)+b_{z},
\end{equation}
\begin{equation}
H_{i}^{GRU}=\left(1-Z_{i}\right) \odot \tilde{H}_{i}^{GCN}+Z_{i} \odot H_{i-1}^{GRU},
\end{equation}
where the parameters and the operator are defined as follows,

		$\odot$ : the product of two matrices on each element.

        $H_i^{GCN}$ : the $i^{th}$ graph of the graph sequence, which is the input of the $i^{th}$ GRU unit.

        $H_{i - 1}^{GRU}$ : the output of previous GRU unit, which is the input of the $i^{th}$ GRU unit. It contains the information of the previous graphs.

        $R_i $ : the output of the reset gate from $H_i^{GCN}$ and $H_{i - 1}^{GRU}$.

		$\tilde{H}_i^{GCN}$ : the output of tanh activation controlled by the reset gate $R_i$, which determines the forgotten degree of the previous state $H_{i - 1}^{GRU}$.

        $Z_i$ : the output of the update gate, which determines how many elements of $H_{i - 1}^{GRU}$ and its state are updated in the $i^{th}$ GRU unit.

        $H_i^{GRU}$ : the output of the $i^{th}$ GRU unit.

        $W_r$, $V_R$, $W_h$ , $V_h$, $W_z$, and $V_z$ are the learnable parameters of the GRU unit.

Each graph $H_{st}^{GCN}$ in $H_s^{GCN}$ is fed into the $t^{th}$ GRU unit to capture the features among the graphs. $H^{GRU}_s$ is obtained by concatenating all outputs of the GRU units together. 

\subsection{Fusion and Classification}
The heterogeneous graph recurrent neural network is used to extract the spatial-temporal domain and spatial-spectral domain features of multi-modal data in the two streams, respectively. After extraction, the features obtained from the spatial-temporal and the spatial-spectral streams are fused, which are defined as: \begin{equation}
H_{3} = (H^{GRU}_{1}W_{1} + b_{1}) \parallel (H_{2}^{GRU}W_{2} + b_{2}),
\end{equation}\begin{equation}
Y = softmax(ReLU(H_3 W_3 + b_3) W_4 + b_4),
\end{equation}
where $\parallel$ denotes the concatenate operation, $H^{GRU}_{1}$ and $H^{GRU}_{2}$ denote the the features extracted by spatial-temporal and spatial-spectral stream, and $Y$ denotes the classification result of our model. $W_{1}$, $b_{1}$, $W_{2}$, $b_{2}$, $W_3$, $b_3$, $W_4$, and $b_4$ denote the learnable parameters of the classification layer. In this paper, the cross entropy \cite{de2005tutorial} is applied as loss function.

\section{EXPERIMENTS}
\subsection{Datasets}
We conduct experiments on two datasets that contain the multi-modal physiological signals: DEAP \cite{koelstra2011deap} and the MAHNOB-HCI \cite{soleymani2011multimodal} datasets, which evoke emotion by multimedia materials.

\textbf{DEAP dataset} records the data generated by 32 participants under multimedia stimulation. Each participant needs to undergo 40 trials and watches a 1-minute music video during each trial. Their physiological signals are recorded while they are watching music videos, and data of each trial includes 3s pre-trial signals and 60s trial signals. The dataset contains 32-channel EEG signals and 8-channel Peripheral Physiological Signals (PPS). The peripheral physiological signals include EOG, EMG, GSR, BVP, respiration, and temperature. The music videos are rated on valence, arousal, and other emotional dimensions from 1 to 9 by all participants.

\textbf{MAHNOB-HCI dataset} records the data generated by 27 participants under multimedia stimulation. Each participant watches 20 video clips, while physiological data of 20 trials is recorded. The length of these video clips is between 34.9s and 117s (mean is 81.4s, standard deviation is 22.5s). The dataset contains 32-channel EEG signals and 6-channel PPS. The peripheral physiological signals include ECG, GSR, respiration, and temperature. Participants are required to score each video clip from 1 to 9 on the emotional dimensions of valence, arousal, etc.

\subsection{Baseline Models and Settings}
The baseline models include MLP \cite{rumelhart1985learning}, SVM \cite{boser1992training}, GCN \cite{kipf2016semi}, DGCNN \cite{song2018eeg}, MM-ResLSTM \cite{ma2019emotion}, ACRNN \cite{tao2020eeg}, and SST-EmotionNet \cite{jia2020sst}. In order to make a fair comparison with the baseline methods, we perform the same data processing and experimental settings for all methods. Specifically, for the DEAP dataset, trials are applied the baseline reduction according to \cite{yang2018emotion}. Then, a 1s non-overlapping window \cite{wang2014emotional} is employed to divide each trial into several samples. For each sample, the Fourier transform with Hanning is utilized to extract the DE features \cite{zheng2015investigating,duan2013differential}. We compute DE features in four frequency bands for each channel. Due to the different attributes of different modal signals, we have adopted different frequency band division strategies for different modal signals. The specific division strategy is detailed in Table S.2 in the Supplementary Material. For the MAHNOB-HCI dataset, similar to the processing of the DEAP dataset, we employ the baseline reduction to each trial and extract the DE features in four frequency bands.

Our model is implemented with Pytorch framework and trained on NVIDIA 2080 Ti. The code has been released in GitHub\footnote{https://github.com/ziyujia/HetEmotionNet}. The threshold is set as 5 to divide valence and arousal into two categories, respectively \cite{koelstra2011deap, tao2020eeg}. Valence measures the pleasantness of emotion and arousal measures the intensity of emotion. We evaluate all methods using 10-fold cross-validation \cite{golub1979generalized}.
\subsection{Results Analysis and Comparison}
We compare our model with other baseline models on the DEAP dataset and the MAHNOB-HCI dataset, respectively. 

\begin{table}[htbp]
\setlength{\abovecaptionskip}{0cm}
\setlength{\belowcaptionskip}{-0.3cm}
\caption{The performance on the DEAP dataset.}
\label{DEAP_result_table}
\begin{tabular}{ccccc}
\toprule
\toprule
Model & Valence (\%)  & Arousal (\%)\\
\midrule
MLP \cite{rumelhart1985learning}    & 74.31$\pm$4.56    & 76.23$\pm$5.12    \\
SVM \cite{boser1992training}        & 83.14$\pm$3.66    & 84.50$\pm$4.43    \\
GCN \cite{kipf2016semi}             & 89.17$\pm$2.90    & 90.33$\pm$3.59    \\
DGCNN \cite{song2018eeg}            & 90.44$\pm$3.01    & 91.70$\pm$3.46    \\
MM-ResLSTM \cite{ma2019emotion}     & 92.30$\pm$1.55    & 92.87$\pm$2.11    \\
ACRNN \cite{tao2020eeg}             & 93.72$\pm$3.21    & 93.38$\pm$3.73    \\
SST-EmotionNet \cite{jia2020sst}    & 95.54$\pm$2.54    & 95.97$\pm$2.86    \\
\midrule
HetEmotionNet   & \textbf{97.66$\pm$1.54}   & \textbf{97.30$\pm$1.65}               \\
\bottomrule
\bottomrule
\end{tabular}
\end{table}

\begin{table}[htbp]
\setlength{\abovecaptionskip}{-0cm}
\setlength{\belowcaptionskip}{-0.5cm}
\caption{The performance on the MAHNOB-HCI dataset.}
\label{MAHNOB-HCI_result_table}
\begin{tabular}{ccccc}
\toprule
\toprule
Model & Valence (\%)  & Arousal (\%)\\
\midrule
MLP \cite{rumelhart1985learning}    & 72.84$\pm$6.88    & 73.61$\pm$10.75    \\
SVM \cite{boser1992training}        & 77.31$\pm$6.77    & 77.16$\pm$9.14    \\
GCN \cite{kipf2016semi}             & 81.31$\pm$6.09    & 83.43$\pm$7.50    \\
DGCNN \cite{song2018eeg}            & 86.21$\pm$9.35    & 86.07$\pm$10.53    \\
MM-ResLSTM \cite{ma2019emotion}     & 89.46$\pm$8.64    & 89.66$\pm$8.09    \\
ACRNN \cite{tao2020eeg}             & 88.10$\pm$5.49    & 89.90$\pm$5.96    \\
SST-EmotionNet \cite{jia2020sst}    & 90.06$\pm$4.80    & 88.37$\pm$7.19    \\
\midrule
HetEmotionNet   & \textbf{93.95$\pm$3.38}   & \textbf{93.90$\pm$3.04}               \\
\bottomrule
\bottomrule
\end{tabular}
\end{table}

Table \ref{DEAP_result_table} presents the average accuracy and standard deviation of these models for emotion recognition on the DEAP dataset. The results indicate our model achieves the best performance on the DEAP dataset. Due to the advantages of automatic feature extraction, most deep learning models have achieved better performance than traditional methods (MLP and SVM). GCN extracts the spatial domain features from signals based on a fixed graph, while DGCNN can adaptively learn the intrinsic relationship of different channels by optimizing the adjacency matrix. Consequently, the performance of DGCNN is more accurate than GCN by around 1\%. MM-ResLSTM uses deep LSTM with residual network to capture the temporal information in the multi-modal signals (EEG and PPS). Hence, MM-ResLSTM can extract more abundant features and perform better than GCN and DGCNN. Compared with MM-ResLSTM, ACRNN can capture both temporal domain and spatial domain information and achieve a better accuracy of 93.72\%, 93.38\% for valence and arousal. Further, SST-EmotionNet can extract the spatial-spectral-temporal domain information from EEG signals by 3D CNN, with the accuracy of 95.54\% and 95.97\%. Although SST-EmotionNet achieves relatively excellent performance, it does not utilize complementarity of multi-modal signals efficiently. Our proposed graph-based model can explore spatial-spectral-temporal domain information from multi-modal signals efficiently. Moreover, our model takes the heterogeneous information of different modalities into consideration. Therefore, our model can adequately learn comprehensive information for emotion recognition, achieving the highest accuracy on the DEAP dataset. Meanwhile, our model achieves the lowest standard deviation, which indicates the stability of our model is also satisfied. Besides, Table \ref{MAHNOB-HCI_result_table} presents the performance of all models on the MAHNOB-HCI dataset. Our model achieves the accuracy of 93.95\% and 93.90\% for valence and arousal, respectively. The performance of other baseline methods is between 72.84\% and 90.06\%. 
The results indicate that our model performs best on the MAHNOB-HCI dataset.

\subsection{Ablation Studies}
To verify the effectiveness of our model, ablation experiments are performed on the DEAP dataset. There are three kinds of ablation experiments: 1$)$ Ablation experiments on the effectiveness of graph transformer network, graph convolutional network, and gated recurrent unit. 2$)$ Ablation experiments on the effectiveness of fusing different modalities. 3$)$ Ablation experiments on two-stream structure to verify whether utilizing spatial-spectral-temporal domain features simultaneously is useful. 

\begin{figure}
\setlength{\abovecaptionskip}{0cm}
\setlength{\belowcaptionskip}{-0.5cm}
\centering
\includegraphics[width=1\linewidth]{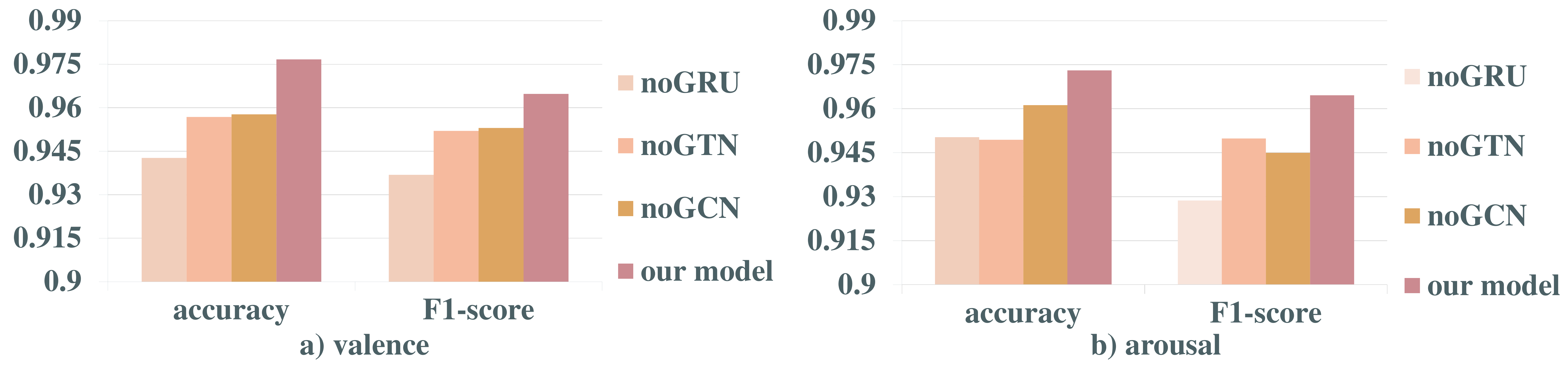}
\caption{Ablation studies on different components.}
\label{ablation_component}
\end{figure}

\textbf{Ablation on different components.}
To verify the effectiveness of different components in our model, we design three variants, which are described as follows:

$\bullet$ HetEmotionNet-noGRU: To verify the effects of GRU on capturing temporal or spectral dependency, this variant removes GRU from HetEmotionNet.

$\bullet$ HetEmotionNet-noGTN: To verify the effectiveness of GTN on modeling the heterogeneity of multi-modal data, this variant removes GTN from HetEmotionNet.

$\bullet$ HetEmotionNet-noGCN: To verify the effectiveness of GCN on capturing the correlation, we remove GCN from HetEmotionNet.

Figure \ref{ablation_component} presents that removing different components from HetEmotionNet reduces the performance. HetEmotionNet outperforms HetEmotionNet-noGTN, which indicates that GTN is effective to model the heterogeneity of multi-modal data. HetEmotionNet has a better performance than HetEmotionNet-noGCN, demonstrating that GCN is effective to capture the correlation among channels. By comparing HetEmotionNet and HetEmotionNet-noGRU, it presents that HetEmotionNet-noGRU performs worse than HetEmotionNet, which indicates that GRU is important to capture the temporal or spectral dependency and improves the performance of model.

\begin{figure}
\setlength{\abovecaptionskip}{0cm}
\setlength{\belowcaptionskip}{-0.5cm}
\centering
\includegraphics[width=1\linewidth]{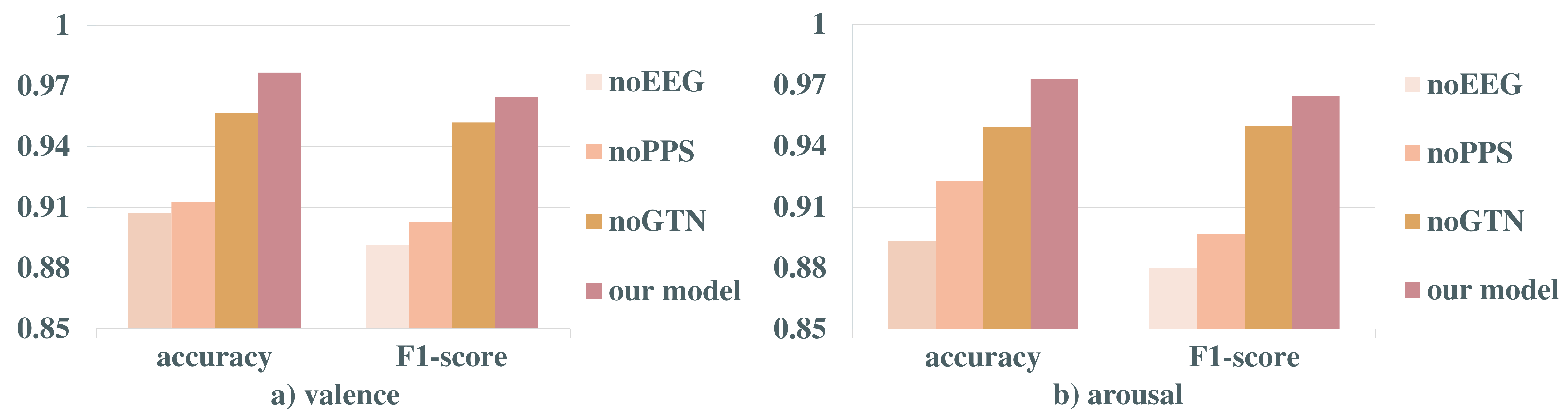}
\caption{Ablation studies on fusing different modalities.}
\label{ablation_multi-modal}
\end{figure}

\textbf{Ablation on fusing data of different modalities.}
To verify the effects of using multi-modal data and modeling the heterogeneity of multi-modal data, we conduct ablation experiments and design three variants of HetEmotionNet:

$\bullet$ HetEmotionNet-noEEG: It removes EEG signals from the multi-modal data to verify the effects of only using PPS on the model's performance. In addition, we remove GTN from this variant because GTN is used to model the heterogeneity of multi-modal data while this variant only uses PPS.

$\bullet$ HetEmotionNet-noPPS: It removes PPS from the multi-modal data to verify the influences of only using EEG signals on the model's performance. We remove the GTN for the same reason of HetEmotionNet-noEEG. 

$\bullet$ HetEmotionNet-noGTN: It uses both EEG signals and PPS. However, it removes GTN to verify that modeling the heterogeneity of multi-modal data is important.

As Figure \ref{ablation_multi-modal} illustrates, HetEmotionNet-noPPS performs better than HetEmotionNet-noEEG. This result indicates that the effects of only using EEG signals is better than only using PPS, because EEG signals are the main physiological signals used in emotion recognition \cite{zhang2020emotion}. Besides, HetEmotionNet-noGTN outperforms HetEmotionNet-noPPS and HetEmotionNet-noEEG, which demonstrates that fusing data of different modalities further improves the performance. However, HetEmotionNet-noGTN does not model the heterogeneity of multi-modal data. Therefore, HetEmotionNet is designed to model the heterogeneity of multi-modal data and reaches the best results. 

\begin{figure}
\setlength{\abovecaptionskip}{0cm}
\setlength{\belowcaptionskip}{-0.66cm}
\centering
\includegraphics[width=1\linewidth]{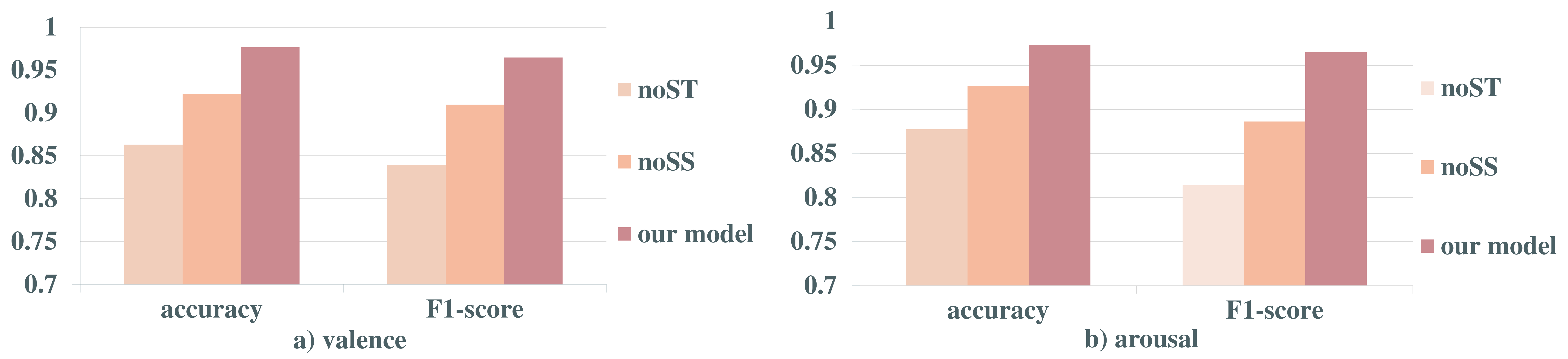}
\caption{Ablation studies on two-stream structure.}
\label{ablation_SST}
\end{figure}

\textbf{Ablation on two-stream structure.}
To verify the effectiveness of integrating two-stream, we design two variants as follows:

$\bullet$ HetEmotionNet-noST (Spatial-Temporal stream): This variant removes the spatial-temporal stream to verify the effects of utilizing the spatial-temporal domain features of physiological signals.

$\bullet$ HetEmotionNet-noSS (Spatial-Spectral stream): This variant removes the spatial-spectral stream to verify the effects of using the spatial-spectral domain features of physiological signals.

As Figure \ref{ablation_SST} illustrates, HetEmotionNet-noSS performs better than HetEmotionNet-noST, which demonstrates that extracting features in spatial and temporal domain is more effective than extracting features in spatial and spectral domain. Besides, HetEmotionNet has a better performance than two variants, which manifests that our model is able to fuse the spatial-spectral-temporal domain features simultaneously and improve the classification accuracy.

\section{CONCLUSION}
In this paper, we propose a novel two-stream heterogeneous graph recurrent neural network using multi-modal physiological signals for emotion recognition. The proposed HetEmotionNet is based on the two-stream structure, which can extract spatial-spectral-temporal domain features from multi-modal signals simultaneously. Moreover, each stream consists of GTN for modeling the heterogeneity, GCN for modeling the correlation, and GRU for capturing the temporal domain or spectral domain dependency. Experiments on the DEAP and the MAHNOB-HCI datasets indicate that our proposed model achieves state-of-the-art performance on DEAP and MAHNOB-HCI. Besides, the proposed model is a general framework for multi-modal physiological signals.

\section*{ACKNOWLEDGMENTS}
Financial supports by National Natural Science Foundation of China (61603029), the Fundamental Research Funds for the Central Universities (2020YJS025), and the China Scholarship Council (202007090056) are gratefully acknowledge. We are grateful for supporting from Swarma-Kaifeng Workshop which is sponsored by Swarma Club and Kaifeng Foundation.

\bibliography{ref}
\balance
\include{supplementary_material}

\end{document}

%% file: supplementary_material.tex
\setcounter{table}{0}
\renewcommand \thetable{S.\arabic{table}}
\onecolumn
\begin{center}\Huge Supplementary Material\end{center}

\begin{table*}[h]
\caption{Notations and explanations}
\label{notation_table}
\begin{tabular}{cc}
\toprule
Notation & Explanation\\
\midrule
$X^F$ & Node features \\
$A$ & Adjacency matrix \\
$G^{ER}$ & Heterogeneous emotional network \\
$T$ & Number of timesteps \\
$B$ & Number of frequency bands \\
$G^T$ & Spatial-temporal graph sequence \\
$G^B$ & Spatial-spectral graph sequence \\
$X^T_u$ & Signals of channel $u$ \\
$a_{u,v}$ & Correlation between channel pair $(u,v)$ \\
$\mathcal{A}$ & Homogeneous adjacency matrix set \\
$Q_l$ & Graph struture generated in the $l^{th}$ GT-layer \\
$L$ & Laplacian matrix \\
$D$ & Degree matrix \\
$U$ & Fourier basis \\
$g_\theta$ & Graph convolution kernel \\
$K$ & Order of Chebyshev Polynomials \\
$\sigma$ & Activation function \\
$ns$ & Number of graphs in a graph sequence \\
$C_A$ & Number of channels \\
$R$ & Output of the reset gate \\
$Z$ & Output of the update gate \\
$W$ & Weight vector \\
$b$ & Bias vector \\
$\phi$ & $1 \times 1$ convolution \\
$\Phi$ & Multi-channel $1 \times 1$ convolution \\
$\odot$ & Element-wise multiplication \\
\bottomrule
\end{tabular}
\end{table*}

\begin{table*}[h]
\caption{The division strategy of frequency bands for physiological signals in different modalities according to [14,34]. For different modalities, the Fourier transform with Hanning is utilized to extract the DE features in four different frequency bands.}
\label{frequency_bands_division}
\begin{tabular}{ccccc}
\toprule
Modality & Frequency band 1 & Frequency band 2 & Frequency band 3 & Frequency band 4 \\
\midrule
EEG & 4-8 & 8-14 & 14-31 & 31-45 \\
EOG & 4-8 & 8-14 & 14-31 & 31-45 \\
EMG & 4-8 & 8-14 & 14-31 & 31-45 \\
ECG & 4-8 & 8-14 & 14-31 & 31-45 \\
GSR & 0-0.6 & 0.6-1.2 & 1.2-1.8 & 1.8-2.4 \\
BVP & 0-0.1 & 0.1-0.2 & 0.2-0.3 & 0.3-0.4 \\
Respiration & 0-0.6 & 0.6-1.2 & 1.2-1.8 & 1.8-2.4 \\
Temperature & 0-0.05 & 0.05-0.1 & 0.1-0.15 & 0.15-0.2 \\
\bottomrule
\end{tabular}
\end{table*}